\documentclass[final,1p,times]{elsarticle}
\usepackage{graphicx}
\usepackage{amssymb}
\usepackage{amsmath}
\usepackage{algpseudocode}
\usepackage{algorithm}
\usepackage{todonotes}
\usepackage{rotating}
\usepackage{multirow}
\usepackage{array}
\usepackage{tabto}
\usepackage{color, soul}
\usepackage{url}
\usepackage[colorlinks]{hyperref}
\usepackage{cleveref}

\begin{document}
\begin{frontmatter}

%\author{
 % Isa Inuwa-Dutse \\
 % School of Engineering and Computer Science\\
 % University of Hertfordshire, United Kingdom\\
 %   i.inuwa-dutse@herts.ac.uk
%}
%\firstnote{Corresponding author:\url{i.inuwa-dutse@herts.ac.uk}} 

%\title{First large collection of diverse Hausa language datasets}
\title{The first large scale collection of diverse Hausa language datasets}
%\title{HauBERT - A pretrained collection of Hausa corpus}

\author{Isa Inuwa-Dutse \\ %Author2 Name2 \\
School of Computer Science\\
University of St Andrews, UK\\
\textit{iid1@st-andrews.ac.uk}} % author2_email_email2

\begin{abstract}
Hausa language belongs to the Afroasiatic phylum, and with more first-language speakers than any other sub-Saharan  African language. With a majority of its speakers residing in the Northern and Southern areas of Nigeria and the Republic of Niger, respectively, it is estimated that over 100 million people speak the language. Hence, making it one of the most spoken Chadic language. While Hausa is considered well-studied and documented language among the sub-Saharan African languages, it is viewed as a low resource language from the perspective of natural language processing (NLP) due to limited resources to utilise in NLP-related tasks. This is common to most languages in Africa; thus, it is crucial to enrich such languages with resources that will support and speed the pace of conducting various downstream tasks to meet the demand of the modern society. While there exist useful datasets, notably from news sites and religious texts, more diversity is needed in the corpus. %In line with the existing studies, the study contributes a diverse Hausa corpus that will significantly help in reducing the amount of time in building tool-kits for various NLP downstream tasks in the language. %low resource language. 

We provide an expansive collection of curated datasets consisting of both formal and informal forms of the language from refutable websites and online social media networks, respectively. The collection is large and more diverse than the existing corpora by providing the first and largest set of Hausa social media data posts to capture the peculiarities in the language. The collection also consists of a parallel dataset, which can be used for tasks such as machine translation with applications in areas such as the detection of spurious or inciteful online content. We describe the curation process -- from the collection, preprocessing and how to obtain the data -- and proffer some research problems that could be addressed using the data. 
\end{abstract}
\end{frontmatter}
%\maketitle

\section{Introduction}
\label{sec:introduction}
\begin{figure}[!tb]
        \centering
        \includegraphics[width=\textwidth]{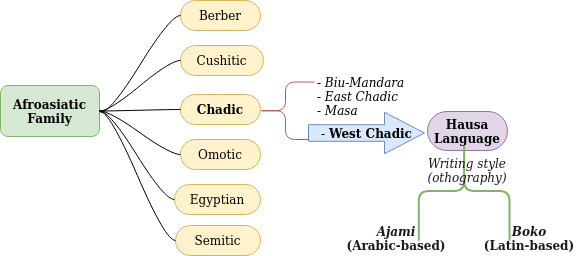}
        \vspace{-8mm}
        \caption{Linguistic family and orthographic origin of Hausa language. The \textit{ajami} is a form of written Hausa text that is based on Arabic alphabets and the \textit{boko} script is based on Latin alphabets.} %the phylogenetic evolution of Hausa language. The orth ....
        \label{fig:hausa-root}
    \end{figure}

Hausa language is one of the most widely spoken Chadic language, which is part of the Afroasiatic phylum. With its majority speakers residing in the Northern and Southern areas of Nigeria and the Republic of Niger, respectively, it is estimated that over 100 million people speak the language. Many countries in West Africa and sub-Saharan countries adopt Hausa as the \textit{lingua franca}, hence making it one of the dominant Chadic language spoken across the continent \cite{jaggar2001hausa}. Major Hausa regions in Nigeria include Kano, Sokoto, Jigawa, Zaria, and Katsina. Figure~\ref{fig:hausa-root} shows a visual depiction of the phylogenetic evolution of Hausa language. The figure shows the linguistic family and orthographic origin of the language. Hausa language relies on Arabic and Latin alphabets (both consonants and vowels). 
\textit{Ajami} is the \textit{Arabic} version of written Hausa that has been widely in existence in pre-colonisation Hausa land, and the \textit{Boko} script is based on \textit{Latin} alphabet, which has been popularised since the down of colonisation in Hausa land. %SEE TABLE XXX FOR CONSONANTS IN BOTH CASES ... 

\paragraph{Motivation} 
In the quest to make computers more amenable to natural language through useful computational tools, relevant datasets are being collected to support various downstream tasks. 
As noted earlier, Hausa is considered a well-studied and documented language among the sub-Saharan African languages \cite{jaggar2001hausa}, but a low resource language from NLP perspective due to limited resources to undertake NLP-related tasks. Because this is common to most languages in Africa \cite{orife2020masakhane}, it is crucial to enrich such languages with resources that will support and speed the pace of conducting various downstream tasks to meet the demand of the modern society. 
Although Google LLC\footnote{\url{https://www.google.com/}} has access to tones of diverse data, anecdotal evidence suggests that its transcription system embedded in the video-communication (\textit{Meet}\footnote{\url{https://meet.google.com/}}) performs poorly in recognising Hausa speech.  

\begin{figure}[t]
        \centering
        \includegraphics[width=\textwidth]{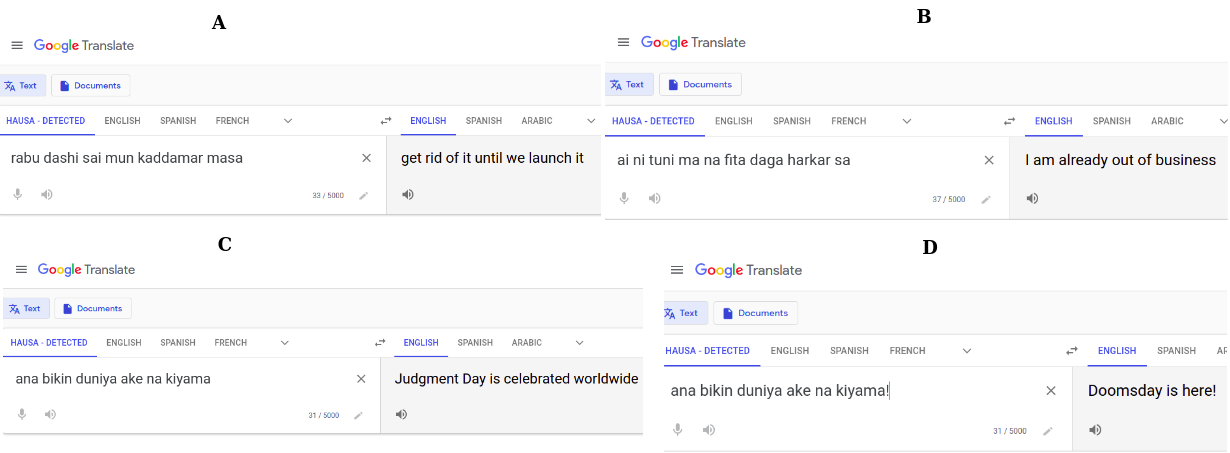}
        \vspace{-8mm}
        \caption{Some examples of translated Hausa texts using Google's translation engine. The texts in \textit{panels A, B, C} and \textit{D} have been incorrectly translated. The use of an exclamation mark (!), \textit{motsin rai} in Hausa, in \textit{panel D} drastically changed the message.}
        \label{fig:text-translation1}
    \end{figure}

\begin{figure}[t]
        \centering
        \includegraphics[width=\textwidth]{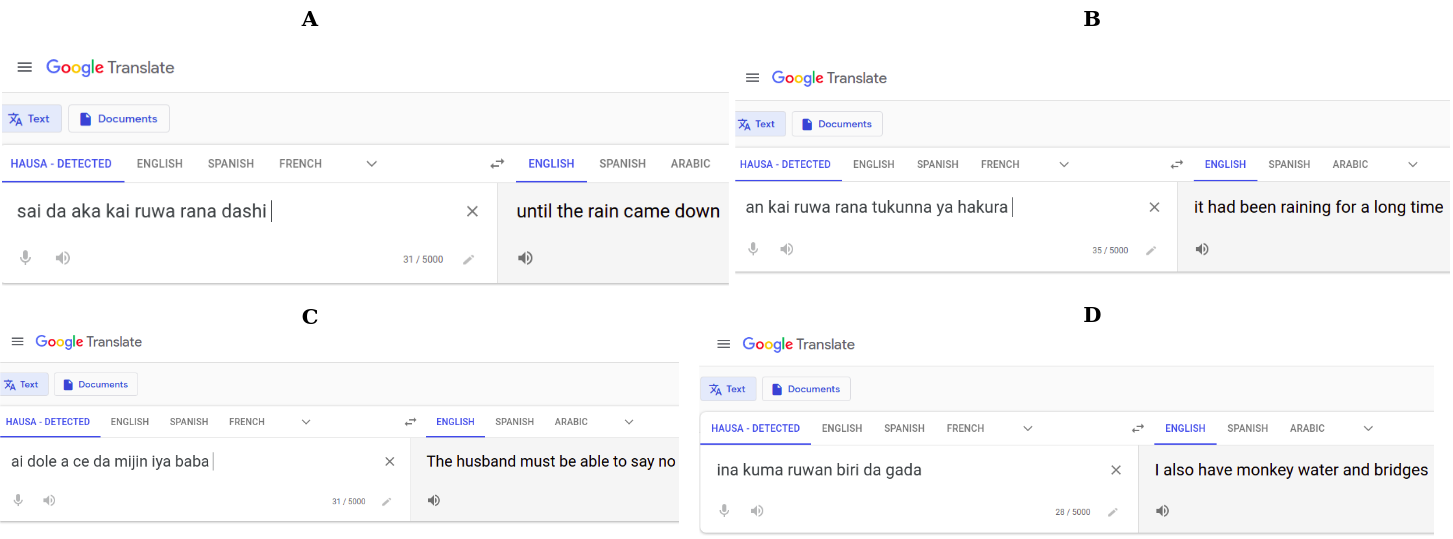}
        \vspace{-8mm}
        \caption{More examples of Hausa texts that have been incorrectly translated. The texts, mostly proverbial, in \textit{panels A, B, C} and \textit{D} are examples of common conversational setting.}
        \label{fig:text-translation2}
    \end{figure}

For a basic assessment of the efficacy of Google's translation engine\footnote{See \url{https://translate.google.co.uk/}} on Hausa language, Figures~\ref{fig:text-translation1} and \ref{fig:text-translation2} show some examples of translated Hausa texts using the translation system. %(see the figure for sample transcripts and the corresponding correct speech). \paragraph{How good is the existing translation system?} %On Hausa Corpus: HOW GOOD IS GOOGLE MEET? See the following for illustrations:
%Hausa: Dazu dazu nan ina cikin tafiya sai na tarar da wani mutum yana gudu %Google’s Meet translation: %Corrected translation: I just saw/encountered a man jogging/running while I was taking a walk 
The discrepancies in the figures (\ref{fig:text-translation1} and \ref{fig:text-translation2}) can be attributed to the lack of sufficient and appropriate data to train relevant language models. Some of the informal or day to day terms used in the language can only be obtained via online social media, not news websites or religious texts. 
%See the translation at \url{https://translate.google.co.uk/?sl=auto&tl=en&text=ana\%20bikin\%20duniya\%20ake\%20na\%20kiyama!&op=translate}
Among the implications of the incorrect translation is misinterpretation or out of context interpretation of events. One of the reasons of the above challenges is the lack of diverse datasets from various sources in larger quantities that will enable the development of relevant resources and tools with a wide range of applications in uncovering meanings in data. The following constitutes the motivations for undertaking this study:
     \begin{itemize}
        \item[-] the existing NLP resources for downstream tasks in Hausa language relies mostly on formal text collections notably from news sites, such as BBC Hausa Service\footnote{\url{https://www.bbc.com/hausa/}} and VOA Hausa Service\footnote{\url{https://www.voahausa.com/}}. Others include parallel datasets from religious texts such as \cite{agic2020jw300} and \textit{Tanzil}\footnote{\url{http://tanzil.net/docs/tanzil_project}} for tasks such as automatic speech recognition, named entity recognition (NER) and part of speech tagging. Thus, the existing Hausa corpora suffer from limited diversity and availability of curated ground-truth data for large scale downstream tasks.  
        \item[-] no known existing collection of curated Hausa datasets from social media platforms, hence, the need for diverse datasets to support various NLP-related tasks.
        \item[-] Other languages have received substantial contributions when compared to the most dominant Chadic language (Hausa). For instance, the HLT for African Languages \cite{adegbola2009building} is focused on notably Yoruba and Igbo. 
     \end{itemize}
This paper aims at addressing the aforementioned gap by making available relevant datasets that will facilitate downstream tasks.

\paragraph{Contributions}
The main focus of this study is to contribute diverse datasets to support NLP research that relies on Hausa language. Noting the above issues, the study offers the following contributions: 
    \begin{itemize}
        \item[-] to enrich the language with useful resources that will facilitate downstream tasks, we provide an expansive collection of curated datasets\footnote{See \url{https://github.com/ijdutse/hausa-corpus} for details.} consisting of both formal and informal forms of the language from refutable websites and online social media networks, respectively. Based on the collection sources, the corpus consists of two broad categories: \textit{Social Stream Category} (Section~\ref{sec:social-stream-category}) and \textit{Website and Blog Category} (Section~\ref{sec:website-category}). The \textit{Social Stream} category consists of Twitter data from users interacting in Hausa language, and the \textit{Website and Blog} category is obtained through the scrapping of relevant websites and blogs. The collection also consists of parallel data, which can be used for natural language processing and machine translation tasks with a wide area of applications.  
        \item[-] because the data is from various sources -- news and government web sites, blog posts and social media platforms -- the collection spans many aspects of social life. Moreover, the inclusion of social media data will help in capturing the peculiarities in the language and how colloquial expressions manifest in the language. 
        \item[-] we make available additional parallel dataset to support downstream tasks such as machine translation and automatic speech recognition systems.
        \item[-] we present a process to streamline the process of obtaining more diverse datasets in Hausa from social media platforms. We describe how to retrieve the hydrated version of the data and proffer some research problems that could be addressed using the data. 
    \end{itemize} 

The remaining of this paper is structured as follows. Section~\ref{sec:related-work} presents a summary of relevant studies and Section~\ref{sec:methodology} presents a detailed description of the data collection and processing. Section~\ref{sec:conclusion} concludes the study and discusses some future work. 
%TABLE XXX FOR ALPHABETS: \textbf{Arabic-based alphabet} (for \textit{ajami} scripts): this consists of 23 characters of the Arabic alphabet () %\textbf{Latin-based alphabet} (for \textit{boko} scripts): this consists of 22 characters of the English alphabet (A/a, B/b, C/c, D/d, E/e, F/f, G/g, H/h, I/i, J/j, K/k, L/l, M/m, N/n, O/o, R/r, S/s, T/t, U/u, W/w, Y/y, Z/z)
% ... the data is available for non-commercial use....such as education/literature, sports, entertainment, religion, banter, colloquial and slang's in the language, perhaps rumours if probe properly, finance, security etc ….

\section{Related Work}
\label{sec:related-work}
As the major spoken language among the sub-Saharan African language, Hausa language is estimated to have over 100 million speakers.  Geographically, the major dialects of the language include Eastern Hausa (e.g. Kano), Western Hausa (e.g. Sokoto) and the Aderanci dialects from Niger. Hausa is considered a well-studied and documented language among the sub-Saharan African languages \cite{jaggar2001hausa}, however, from NLP perspective, it is viewed as a low resource language due to limited resources for computational linguistic. This is common to most languages in Africa. There is a growing body of studies exploring various aspects of the language. Of interest to this study is the prevailing efforts to enrich the language along the following dimensions. 

\paragraph{Existing Corpus} 
From a computational linguistic point of view, Hausa is categorised as low-resource language owing to the limited resources to computationally analyse the language and support downstream tasks. Other languages have received substantial contribution when compared to the most dominant Chadic language (Hausa) widely spoken in Africa. Generally, there are limited linguistic resources to support NLP tasks involving African languages. The commonly used parallel datasets for NLP tasks, such as Machine Translation, include (1) Tanzil dataset\footnote{\url{http://tanzil.net/docs/tanzil_project}}, which is a translation of the Holy Qur'an in various languages including Hausa. %(127k parallel sentences and less than 10k unique Hausa sentences)
(2) JW300 dataset \cite{agic2020jw300} consisting of multilingual parallel sentences, and is used in the \textit{Masakhane} Project \cite{orife2020masakhane} as the default dataset for open implementation.  
So far, the \textit{Masakhane} Project\footnote{\url{https://www.masakhane.io/}} is one of the biggest collaborative attempts to support African language developments through the development and support for open-source linguistic resources and tools for African languages \cite{orife2020masakhane}. The work of \cite{akinfaderin2020hausamt} noted the lack of adequate parallel and diverse dataset for language translation. This led the authors to embark on a project aimed at improving machine translation capability for low-resource language. The data consists of \textit{Hausa–English} parallel corpus, which is useful for translation and as evaluation benchmark for \textit{English–Hausa} Neural Machine Translation (NMT). 

\paragraph{Language Models}
The work of \cite{abdulmumin2019hauwe} contributes a Hausa-based words embedding model that is trained on formalised texts from news web sites and religious texts to support downstream tasks. Noting the inadequacy of models trained using high-resource languages (such as English) to be transferred for training low resource languages (such as Hausa), the work of \cite{hedderich2020transfer} is based on multilingual transformer models to improve accuracy and effectiveness of low resource language models. The use of more diverse datasets from various sources such as social networks will help in capturing the peculiarities in the language, thus improving the language models.

\paragraph{Downstream tasks in low resource languages}
There are relevant parallel datasets in place to support downstream tasks such as machine translation and automatic speech recognition (ASR) systems. Noting how the limited resources available to facilitate Human Language Technology (HLT), the work of \cite{adegbola2009building} is focused on speech technology for some dominant indigenous languages in Nigeria \cite{adegbola2009building}. Similarly, the work of \cite{schlippe2012hausa} presents a database of text and speech, transcribed speech resources, in Hausa language. The speech collection is based on the GlobalPhone project (\cite{schultz2002globalphone}), and the textual part collected from Hausa-based websites, mostly from news sites. Using the collected datasets, the authors develop an ASR system to improve pronunciation and acoustic modelling in Hausa language. 

Taking the aforementioned resources into account, the diversity of the datasets utilised can be improved by incorporating various sources that capture more dialects and social media data texts. Thus, the datasets in this study can be used for developing and improving relevant resources for downstream tasks.

\section{Datasets}%Experimentation: Building the Corpus
\begin{figure}[t]
        \centering
        \includegraphics[width=\textwidth]{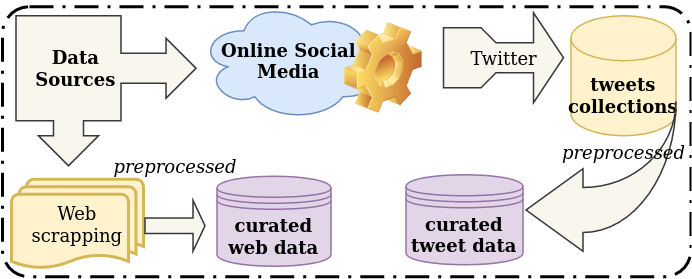}
        \vspace{-8mm}
        \caption{A summary of the data collection pipeline.}
        \label{fig:data-collection-pipeline}
    \end{figure}

\label{sec:methodology} 
In the quest to make computers more amenable to natural language through useful computational tools, numerous collection of diverse datasets are being collected. Based on the collection sources, the corpus is categorised into the following: \textit{Social Stream} and \textit{Website and Blog} categories. Accordingly, the collection consists of two sub-categories of Twitter data, from users interacting in Hausa language, and two sub-categories of web scraping data from various Hausa-based language sites and historic online files. The collection also consists of parallel data, which can be used for downstream tasks such as machine translation. Figure~\ref{fig:data-collection-pipeline}\footnote{the figure is first used in \cite{inuwa2020towards}} shows a summary of the data collection pipeline.
    %\begin{figure}[!tb]

\subsection{Social Stream Category}
\label{sec:social-stream-category}
The proliferation of various online information on social media platforms such as Twitter and Facebook makes it possible to harvest relevant posts. With the evolving nature of communication in which majority of the populace obtain useful information from the web and social media platforms, it is pertinent to retrieve as much as possible relevant online datasets on low resource languages. In view of this, the \textit{Social Stream} collection contains \textit{tweet objects}\footnote{a \textit{tweet object} is a complex data object composed of numerous descriptive fields, which enables the extraction of various features related to a post} retrieved from Twitter using the platform's application programming interface (API). The collection activity starts on 11/12/20 to date (13/02/21) and consists of the following sub-categories. 
    \begin{itemize}
        \item[-] \textbf{account-based collection:} this relies on specific usernames, mostly of news channels, to collect data posted in Hausa language. Because of the presence of news channels, the category usually consists of formal posts on news items that attract other users to engage through \textit{comment, reply, retweet} or \textit{like}. This allows for the retrieval of more data on users' engagements using slang, memes and other colloquial or informal posting styles in the local language. 
        \item[-] \textbf{miscellaneous collection:} the miscellaneous set mostly consists of users who engage or interact with the content posted by users in the \textit{account-based collection}. For a more diverse dataset, the miscellaneous set also consists of post collected using relevant \textit{hashtags} and \textit{keywords} (see Table~\ref{tab:data-sources-summary}). The use of \textit{hashtags} and \textit{keywords} offers a generic collection of daily tweets spanning numerous topics posted in Hausa language.   
    \end{itemize}
We define \textit{main tweet}, as a post or tweet that attracts engagements by other users via replies. For annotation and sharing purpose, each main tweet (prefixed with MT) is followed by its corresponding reply, prefixed with MTR. Under this setting, the main tweet is not unique but the reply is.

\paragraph{Getting the hydrated tweets} 
The full version of the \textit{social stream category} cannot be made available due to Twitter’s policy on content redistribution\footnote{see \url{https://developer.twitter.com/en/developer-terms/agreement-and-policy}}. 
Instead, we provide relevant IDs of the \textit{tweet-object} to be used in retrieving the complete data, including the meta-information about each tweet\footnote{The data presented in this paper are available at \url{https://github.com/ijdutse/hausa-corpus}}. Moreover, we include a short \textit{Jupyter notebook} with a description of how to retrieve the data, especially the hydrated data, and render it usable. Alternatively, interested users may use the \textit{hydrator package} \cite{hydrator2020}. 
   \begin{table}[t]
        \small
        \caption{Summary of data collection sources and corresponding descriptions} 
        \label{tab:data-sources-summary}
        \begin{tabular}{p{2.15cm} p{11cm}}
        \hline
        \textbf{Social Stream}   & \textit{account-based collection} from Twitter   \\
        \textbf{Description}    & tweets from specific account handles collected via Twitter's API \\
        \textbf{Handles}   & 
                @bbchausa, @voahausa, @HausaDw, @RFI\_Ha, @freedomradionig, @aminiyatrust, @HausaRadio, @hausaonline, \ 
                @rariyajarida, @CRIHausaRadio, @TRTHausa, @HausaTranslator, @HausaDictionary, @PTimesHausa, \\ \hline
        \textbf{Social Stream}   & \textit{miscellaneous collection} from Twitter \\ 
        \textbf{Description}    & tweets collected from Twitter using diverse keywords\\
        \textbf{Keywords}   & 
                Hausa language, Yaran Hausa, Dan Hausa, Hausa day, Hausa week, barkan mu da rana, 
                sannu da zuwa, lokaci, gobe da rana, sanyi, yan siyasa, majalisa, shugabanni, labarai,
                \#HausaDay, yan sanda, sojoji, wasan kwallon kafa, \#hausaday \\ \hline
        \textbf{Website \& Blog}   & \textit{Major websites for non-tweet-based collection}    \\ 
        \textbf{Description}    & more formal texts from major news channels and a chronicle of posts\\
        \textbf{Some urls}   & 
                 \url{https://www.bbc.com/hausa, https://www.voahausa.com/, https://freedomradionig.com/, http://indigenousblogs.com/ha/, https://www.rumbunilimi.com.ng/, http://fagenmarubuta.blogspot.com/} 
                 \\ \hline
        \end{tabular}
    \end{table} 

\subsection{Website and Blog Category}
\label{sec:website-category}
This category comprises data from Hausa-based news sources, blog posts and from historic files from various sources. Using a custom web scrapping\footnote{using the \textit{BeautifulSoup} package, available at \url{www.crummy.com/software/BeautifulSoup}} crawler, we scrap useful data from the relevant websites; see Table~\ref{tab:data-sources-summary} for some of the visited sites and Appendix~\ref{sec:appendix} for more details. Because the content in this category mostly consists of well-written texts, it is possible to examine and compare the formal written form of the language with its informal counterpart from social media platforms. 
 
\subsection{Parallel Data} 
\label{sec:parallel-data}
As a first step in creating a parallel corpus (in English) for the corpus (in Hausa), we leverage Google's translation API, \textit{Googletrans}\footnote{\url{https://translate.google.co.uk/}}, to efficiently create a parallel English corpus for the collected Hausa corpus (using the \textit{Social Stream} category). We sample from the \textit{Social Stream} category for the translation, and subset of the translated segments or sentences is manually validated for correctness. Similar to the \textit{Social Stream} category, the parallel collection consists of \textit{Hausa main tweet} (HMT), \textit{Hausa main tweet reply} (HMTR), \textit{English main tweet} (EMT) for the translated HMT, and \textit{English main tweet reply} (EMTR) for the translated HMTR. 

\subsection{Preprocessing and meta-analysis} 
\label{sec:preprocessing}
Due to the multifaceted nature of the datasets, we applied the following transformation so that all the collections conform to a unified format for redistribution. This is to improve usability and suitability for sharing and further analysis. 

\paragraph{Preprocessing} 
Noting how messy the \textit{Social Stream} category could be and difficult to use directly without cleaning \cite{inuwa2018detection}, we applied the following rudimentary cleaning process involving tokenisation, stopwords removal and text formatting. For normalisation, we remove \textit{urls, user mentions, emojis} and \textit{non-Hausa posts} from the collection. Both the raw and the cleaned version of the applicable datasets are made available because some of the raw tweets or replies contain only \textit{user mention} or \textit{emojis}, which we designate as stopwords in the cleaned version of the data. After the removal of spaces and special symbols in the \textit{Website and Blog} category, each paragraph or chunk of texts is split into smaller unit punctuated by a \textit{full stop}. 

\begin{figure}[t]
        \centering
        \includegraphics[width=\textwidth]{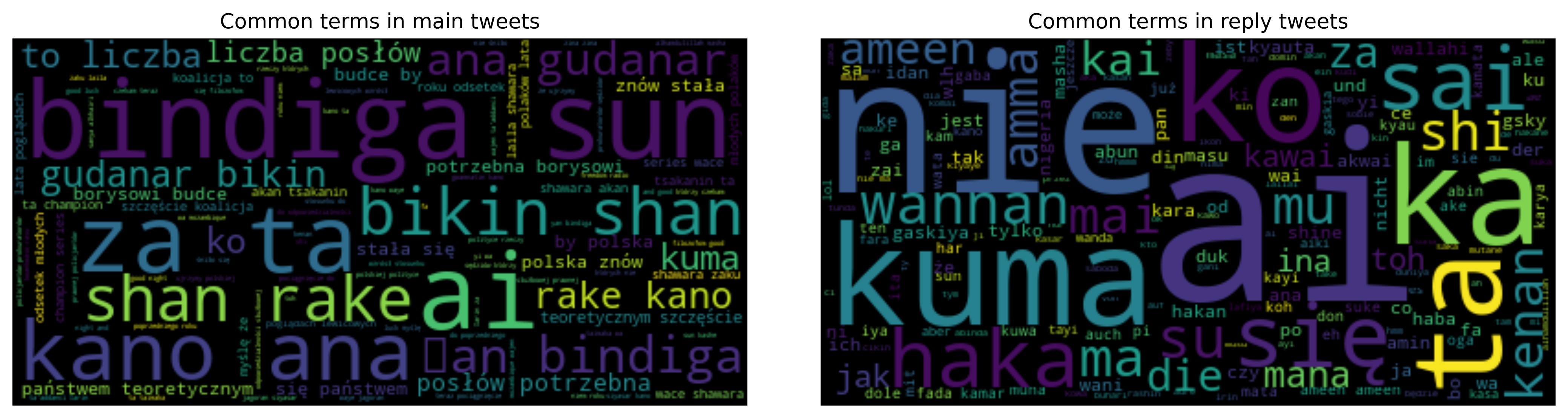}
        \vspace{-8mm}
        \caption{Some common terms in the tweets collection.}
        \label{fig:word-cloud-of-common-terms}
    \end{figure}

\paragraph{Meta-analysis} 
Because the datasets in the corpus have been collected over time, we conducted longitudinal and exploratory analyses for better understanding. Table~\ref{tab:data-statistics} shows basic statistics about the collected data, including the number of unique main tweets, replies, mean token size, and the average length of sentences in each category. %\textbf{Figure xxx} shows a timeline of popular tweets overtime during the collection period in which the \textit{y-axis} denotes the frequency of tweets and the \textit{x-axis} denotes the popular hashtags and the corresponding dates in the title. 
For the \textit{Social Stream} category, we determine the most common terms through a high-level visualisation of samples from the tweets collection. Accordingly, we provide a summary of the relevant themes in Figure~\ref{fig:word-cloud-of-common-terms} using \textit{word cloud}\footnote{the word cloud visualisation is based on the implementation at: \url{https://github.com/amueller/word_cloud}}.
       \begin{table}[t]
              \footnotesize
              \caption[basic stats]{A summary of basic statistics in the collected datasets. \#DS and \#TS denote dataset size and tokens size the collection, respectively; $\mu_{sent.}$ refers to the average sentence size in each data category}
              \label{tab:data-statistics}
              \begin{tabular}{l>{\raggedright}p{3.0cm}ccclp{3cm}} \hline
                & Category &  \#DS & \#TS & $\mu_{sentence~size}$& Description  \\ \hline 
                \multirow{6}{*}{\rotatebox{90}{\textbf{Datasets}}} 
                & Main Tweets & 3389 & 34,035  & 10 & Tweets from social stream  \\
                & Reply Tweets & 8649 & 61,441  & 7 & Tweets from social stream \\
                & News and Blogposts & 29,371 & 932,523  & 32 & Data collection from websites and blogs\\
                & Web Documents & 12,277 & 10507117 & 856 & Data collection from websites and blogs\\
                & Parallel Main Tweets & 18 & 3671 & 13 & Parallel data from social stream \\ 
                & Parallel Reply Tweets & 292 & 3671 & 7 & Parallel data from social stream \\ \hline
                \hline
              \hline
            \end{tabular}
            \normalsize
        \end{table}

     %\begin{figure}[!tb]

\subsection{Utility of the datasets}
\label{sec:data-utility}
Because this is the first large scale diverse collection of datasets in Hausa language, many relevant studies can be conducted using the corpus to complement existing resources. Thus, we identify relevant areas and research problems that could harness the curated datasets.
    \begin{itemize}
        \item[-] \textbf{detection of online spurious content:} the increasing number of uncensored posts is posing many challenges to online socialisation, especially the problem of identifying genuine information in an ecosystem cluttered with spurious content. Despite the prevention measures taken by social media platforms, many sources of misleading information and rumours still exist, especially in low-resource languages such as Hausa. The collected corpus, especially the \textit{Social Stream} category, could be used to develop a prediction system to ascertain content veracity. %and detect spurious and inciting online content
        \item[-] \textbf{propagation of fake news:} noting the prevalence of online fake news, it will be crucial to understand how such fake news and unfounded claims propagate in low resource languages, especially since the existing methods are based on high resource languages. The contributed datasets will be useful in investigating the problem. 
        \item[-] \textbf{low resource language model:} the diverse nature of the corpus will improve the training of useful state-of-the-art language models, such as BERT \cite{devlin2018bert}. While variants of such models have been used for multilingual setting, the data collection used for training is shallow and lacks diversity. Among other applications, the corpus can be used for downstream tasks such as machine translation, automatic speech recognition system, and sentiment analysis in the low resource language.  
    \end{itemize}
Overall, the contribution of this study will significantly help in reducing the amount of time or effort in building tool-kits for downstream tasks in the low resource language. 

\section{Conclusion}
\label{sec:conclusion} 
Hausa language is one of the dominant Chadic languages that is widely spoken in Africa. %Hausa language belongs to the Afroasiatic language phylum, and with more first-language speakers than any other sub-Saharan  African language. With a majority of its speakers residing in the Northern and Southern areas of Nigeria and Republic of Niger, respectively, it 
From the computational linguistic point of view, Hausa, among other African languages, is categorised as low-resource language owing to the limited resources to handle various NLP-related tasks. This limitation slows down the pace of development of the language in terms of computational linguistic and for downstream tasks. 
In line with the existing studies, the study contributes a diverse Hausa corpus that will significantly help in reducing the amount of time in building tool-kits for various NLP downstream tasks in the language. We contributed the first comprehensive collection of curated datasets to support various NLP-related tasks in Hausa language. Essentially, the corpus consists of two categories according to collection sources: online social network and websites and blogs. Due to the collection sources, the themes in the datasets span many aspects of social life. Moreover, the inclusion of social media data will help in capturing the peculiarities in the language and how colloquial expressions manifest in the language. %Which is hitherto unavailable .... %To create a parallel dataset and develop/create language model. The parallel datasets will help in the development of machine translation systems, part-of-speech (POS) tagging (especially involving social media posts), 
We provide a basic but useful analysis of the datasets and recommend some relevant problems that could leverage the corpus. We also presented a process to streamline the activity of obtaining more diverse datasets in Hausa from social media platforms, and how to retrieve the hydrated version of the data. 

\paragraph{Future work} It is pertinent to have a comprehensive collection and streamline the process to harness in enriching the language with the relevant resources. For future study, we hope this will attract research curiosity to further explore the language in-depth for supporting downstream tasks. Future focus will be on the development of a language model using the contributed datasets and other existing ones. The diverse nature of the corpus will improve the training of useful state-of-the-art language models, such as BERT \cite{devlin2018bert}. As pointed earlier in Section~\ref{sec:data-utility}, the corpus can be used for downstream tasks such as machine translation, automatic speech recognition system, and sentiment analysis in the low resource language. Thus, it will be useful to develop a more diverse translation system that is equipped to support the informal Hausa content commonly found in online social media platforms such as Facebook and Twitter. %it will be crucial to understand how such fake news and unfounded claims propagate in low resource languages, especially since the existing methods are based on high resource languages. 

% such as automatic speech recognition (ASR). Despite its accessibility to tones of diverse data, anecdotal evidence suggests that Google's translation API performs poorly in recognising Hausa speech (see attached transcript and corresponding speech).... this technologically-driven society. .... Despite its millions of speakers and active use of digital infrastructure (mostly youths?, but with only very limited support for NLP technologies ... One of the reasons is the lack of diverse datasets from various sources in larger quantities. %dissemination of public health information in low resource translation setting.% .... that will enable the development of relevant resources and tools with a wide range of applications in uncovering meanings in data. % from the low resource language (Hausa). The goal is to enrich the language with useful resources that will facilitate downstream tasks. We describe the data curation process -- from collection to training of a language model suited for the language. Also, we describe how to retrieve the hydrated version of the social media version and proffer some research problems that could be addressed using the data. 
%################################

\section*{Appendix}
\label{sec:appendix}
Details about the visited \textit{urls} used in the \textit{Website and Blog} category (Section~\ref{sec:website-category}) can be found at the project's Github repository available at \url{https://github.com/ijdutse/hausa-corpus/tree/master/data/web-sites}. Other relevant resources include the following: %For more relevant studies see the following: 
    \begin{itemize}
        \item[-] HugginFace: \url{https://huggingface.co/datasets/hausa_voa_ner#curation-rationale}
        \item[-] Published Hausa Corpora: \url{https://github.com/besacier/ALFFA_PUBLIC/tree/master/ASR/HAUSA}
        \item[-] Library of Congress: \url{https://www.loc.gov/rr/pdf/requestingmaterials.pdf}
    \end{itemize}

%\section*{Acknowledgements} 
\bibliographystyle{unsrt}
\bibliography{hausa-corpus}
\end{document}